\documentclass{article}

\usepackage{arxiv}

\usepackage[lithuanian,english]{babel}
\usepackage[utf8]{inputenc} 
\usepackage[T1]{fontenc}    
\usepackage{hyperref}       
\usepackage{url}            
\usepackage{booktabs}       
\usepackage{amsfonts}       
\usepackage{nicefrac}       
\usepackage{microtype}      
\usepackage{multirow}        
\usepackage{graphicx}
\usepackage[square,numbers,sort&compress]{natbib}
\usepackage{doi}

\usepackage{xcolor}         

\title{Comparison of Modern Multilingual Text Embedding Techniques for Hate Speech Detection Task}

\newif\ifuniqueAffiliation
\uniqueAffiliationfalse

\ifuniqueAffiliation 
\author{ \href{https://orcid.org/0000-0002-4769-4527}{\includegraphics[scale=0.06]{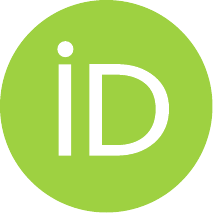}\hspace{1mm}Evaldas Vaičiukynas} \\
	Department of Information Systems\\
	Faculty of Informatics\\
	Kaunas University of Technology\\
	\texttt{evaldas.vaiciukynas@ktu.lt} \\
	\And
	\href{https://orcid.org/0000-0002-2054-0624}{\includegraphics[scale=0.06]{orcid.pdf}\hspace{1mm}Paulius Danėnas} \\
	Centre of Information Systems Design Technologies\\
	Faculty of Informatics\\
	Kaunas University of Technology\\
	\texttt{paulius.danenas@ktu.lt} \\
}
\else
\usepackage{authblk}

\newbox{\orcid}\sbox{\orcid}{\includegraphics[scale=0.06]{orcid.pdf}} 
\author[1]{
	\href{https://orcid.org/0000-0002-4769-4527}{\usebox{\orcid}\hspace{1mm}Evaldas Vaičiukynas\thanks{\texttt{evaldas.vaiciukynas@ktu.lt}}}
}
\author[2]{
	\href{https://orcid.org/0000-0002-2054-0624}{\usebox{\orcid}\hspace{1mm}Paulius Danėnas\thanks{\texttt{paulius.danenas@ktu.lt}}}
}
\author[1]{
	\href{https://orcid.org/0000-0002-5594-4733}{\usebox{\orcid}\hspace{1mm}Linas Ablonskis\thanks{\texttt{linas.ablonskis@ktu.lt}}}
}
\author[1]{
	\href{https://orcid.org/0000-0002-5594-4733}{\usebox{\orcid}\hspace{1mm}Algirdas Šukys\thanks{\texttt{algirdas.sukys@ktu.lt}}}
}
\author[2]{
	\href{https://orcid.org/0000-0001-8546-0564}{\usebox{\orcid}\hspace{1mm}Edgaras Dambrauskas\thanks{\texttt{edgaras.dambrauskas@ktu.lt}}}
}
\author[1]{
	\href{https://orcid.org/0000-0002-0364-5985}{\usebox{\orcid}\hspace{1mm}Voldemaras Žitkus\thanks{\texttt{voldemaras.zitkus@ktu.lt}}}
}
\author[1]{
	\href{https://orcid.org/0000-0003-3250-4599}{\usebox{\orcid}\hspace{1mm}Rita Butkienė\thanks{\texttt{rita.butkiene@ktu.lt}}}
}
\author[1,2]{
	\href{https://orcid.org/0000-0001-6396-2483}{\usebox{\orcid}\hspace{1mm}Rimantas Butleris\thanks{\texttt{rimantas.butleris@ktu.lt}}}
}
\affil[1]{Department of Information Systems, Faculty of Informatics, Kaunas University of Technology, Kaunas, Lithuania}
\affil[2]{Centre of Information Systems Design Technologies, Faculty of Informatics, Kaunas University of Technology, Kaunas, Lithuania}
\fi

\renewcommand{\shorttitle}{Multilingual Text Embeddings for Hate Speech Detection}

\hypersetup{
pdftitle={Multilingual Text Embeddings for Hate Speech Detection},
pdfsubject={Computer Science},
pdfauthor={Vaiciukynas, Danenas, Ablonskis, Sukys, Dambrauskas, Zitkus, Butkiene, Butleris},
pdfkeywords={Hate speech detection, Anomaly detection, Sentence embeddings, Dimensionality reduction},
}

\begin{document}
\maketitle

\begin{abstract}
Online hate speech and abusive language pose a growing challenge for content moderation, especially in multilingual settings and for low-resource languages such as Lithuanian. This paper investigates to what extent modern multilingual sentence embedding models can support accurate hate speech detection in Lithuanian, Russian, and English, and how their performance depends on downstream modeling choices and feature dimensionality. We introduce LtHate, a new Lithuanian hate speech corpus derived from news portals and social networks, and benchmark six modern multilingual encoders (potion, gemma, bge, snow, jina, e5) on LtHate, RuToxic, and EnSuperset using a unified Python pipeline. For each embedding, we train both a one class HBOS anomaly detector and a two class CatBoost classifier, with and without principal component analysis (PCA) compression to 64-dimensional feature vectors. Across all datasets, two class supervised models consistently and substantially outperform one class anomaly detection, with the best configurations achieving up to 80.96\% accuracy and AUC ROC of 0.887 in Lithuanian (jina), 92.19\% accuracy and AUC ROC of 0.978 in Russian (e5), and 77.21\% accuracy and AUC ROC of 0.859 in English (e5 with PCA). PCA compression preserves almost all discriminative power in the supervised setting, while showing some negative impact for the unsupervised anomaly detection case. These results demonstrate how modern multilingual sentence embeddings combined with gradient boosted decision trees provide robust soft-computing solutions for multilingual hate speech detection applications.
\end{abstract}


\section{Introduction}
\label{sec:intro}

The rapid proliferation of digital communication platforms has fundamentally transformed the way societies interact, debate, and share information. While social media, online news portals, and public forums have democratized access to public discourse, they have simultaneously become vectors for the dissemination of hate speech --- language that attacks, demeans, or incites violence against individuals or groups on the basis of protected characteristics such as race, ethnicity, religion, gender, sexual orientation, or national origin \cite{Fortuna2018, Poletto2021}. The scale and urgency of this phenomenon are staggering: a 2023 global survey conducted by UNESCO and Ipsos across 16 countries found that 67\% of internet users have personally encountered hate speech online, with the prevalence rising to 74\% among users under the age of 35 \cite{UNESCO2023}. Similarly, a European Union survey has reported that approximately 80\% of respondents in the EU have encountered hate speech in online environments \cite{GaglioneEtAl2021}. These figures underscore that online hate speech is not a marginal or isolated problem but a pervasive feature of the contemporary digital landscape that affects billions of users worldwide.

The societal consequences of unchecked online hate speech extend far beyond individual psychological harm. At the individual level, exposure to hateful content has been associated with increased anxiety, depression, social withdrawal, and, in extreme cases, suicidal ideation \cite{Hawdon2017, GaglioneEtAl2021}. At the community level, persistent hate speech fosters social polarization, normalizes bigotry, and erodes the quality of public discourse \cite{GaglioneEtAl2021}. More alarmingly, a growing body of empirical research has established direct links between the spread of online hate speech and real-world violence. M\"{u}ller and Schwarz \cite{MullerSchwarz2021} demonstrated that anti-refugee hate speech on Facebook causally predicted violent crimes against refugees in German municipalities, with the effect disappearing during major platform outages. The role of social media in facilitating mass atrocities has been tragically illustrated in Myanmar, where Facebook was identified by a United Nations fact-finding mission as an instrument for those seeking to spread hate against the Rohingya Muslim minority, contributing to a campaign of ethnic cleansing that displaced over 700,000 people \cite{UNMyanmar2018}. In light of such evidence, the need for scalable and reliable automated hate speech detection systems has become a pressing concern for governments, platform operators, and civil society alike.

The regulatory environment has evolved accordingly. The European Union's Digital Services Act (DSA), which became fully applicable in February 2024, imposes explicit obligations on large online platforms to address illegal content --- including hate speech --- through a combination of automated detection, human review, and transparent reporting mechanisms \cite{DSA2022}. In January 2025, the European Commission integrated the revised Code of Conduct on countering illegal hate speech online into the DSA framework, further strengthening the requirements for proactive content moderation. These regulatory developments have intensified the demand for automated content moderation tools that are accurate across linguistic and cultural contexts, scalable to the volume of user-generated content (which exceeds millions of posts per minute on major platforms), and robust against adversarial evasion strategies \cite{Vidgen2020}.

From a natural language processing (NLP) perspective, hate speech detection is typically framed as a supervised text classification problem. The field has evolved rapidly from early approaches based on handcrafted features and classical machine learning algorithms --- such as bag-of-words representations paired with support vector machines or logistic regression \cite{Schmidt2017, Davidson2017} --- to deep learning architectures leveraging word embeddings \cite{Gambaeck2017, Badjatiya2017} and, more recently, pre-trained transformer-based language models \cite{vaswani2017, Mozafari2020a, Caselli2021}. Fine-tuned variants of BERT \cite{Devlin2018} and its multilingual counterpart mBERT, as well as cross-lingual models such as XLM-RoBERTa \cite{journalscorrabs191102116}, have achieved state-of-the-art performance on several English-language hate speech benchmarks and have demonstrated the ability to transfer detection capabilities across languages via shared multilingual representation spaces \cite{zampieri2020, Ranasinghe2020, Roy2020}.

Despite these advances, several critical challenges remain. First, the overwhelming majority of hate speech detection research has concentrated on English-language data, leaving most of the world's approximately 7,000 languages without adequate detection tools or annotated resources \cite{Fortuna2018, Poletto2021, Vidgen2020}. Recent surveys have catalogued over 60 publicly available hate speech training datasets, yet the vast majority are English-centric, and only a handful cover languages from Central and Eastern Europe, the Baltics, or other underrepresented linguistic communities \cite{Vidgen2020, Tonneau2024}. For low-resource languages, the scarcity of annotated corpora, the absence of language-specific NLP pre-processing tools, and the limited coverage of pre-trained models compound the difficulty of building effective detection systems \cite{Bigoulaeva2021, Bigoulaeva2022}. Lithuanian is a prototypical example of such a low-resource scenario: despite the existence of active online communities and documented instances of online hate speech in the Lithuanian digital sphere, systematic studies on automated hate speech detection in Lithuanian have only recently begun to appear in the literature \cite{Mandravickaite2025}.

Second, modern multilingual sentence embedding models have emerged as a promising paradigm for cross-lingual and multilingual text classification tasks. Models such as Multilingual E5 \cite{wang2024multilinguale5textembeddings}, Jina Embeddings \cite{sturua2024jinaembeddingsv3multilingualembeddingstask}, Snowflake Arctic \cite{yu2024arcticembed20multilingualretrieval}, and BGE-M3 \cite{chen2025m3embeddingmultilingualitymultifunctionalitymultigranularity} encode texts from dozens or hundreds of languages into a shared vector space, enabling downstream classifiers to operate on fixed-dimensional representations without requiring language-specific fine-tuning. However, systematic comparisons of these modern off-the-shelf embedding models for hate speech detection --- particularly in multilingual settings that include low-resource languages --- remain scarce. Most existing studies either focus on a single encoder, employ task-specific fine-tuning that obscures the contribution of the base representation, or evaluate only on well-resourced languages. This gap motivates a rigorous, controlled comparison of modern embedding techniques across diverse linguistic settings under a unified experimental protocol.

Third, practical deployment of hate speech detection systems in real-world content moderation pipelines imposes stringent constraints on computational efficiency, memory footprint, and inference latency. Dimensionality reduction techniques, such as Principal Component Analysis (PCA), offer a straightforward mechanism for compressing high-dimensional embeddings while potentially preserving discriminative information. Understanding the trade-off between compression and detection performance is critical for designing systems that can operate at scale without prohibitive computational costs, yet this aspect has received limited systematic attention in the hate speech detection literature.

In this paper, we address these gaps by investigating the effectiveness of several recent multilingual text embedding models for hate speech detection in three languages: Lithuanian, Russian, and English. We focus on sentence-level vector representations produced by six modern multilingual encoders --- Potion, Gemma, BGE, Snowflake Arctic, Jina Embeddings (v3), and Multilingual E5 (large-instruct) --- and evaluate them in combination with a two-class CatBoost supervised classifier and a one-class HBOS anomaly detection approach across three binary hate speech datasets, with and without PCA-based dimensionality reduction.

Our main scientific contributions are as follows:

\begin{itemize}
\item We devise a unified experimental framework for benchmarking multilingual sentence embeddings on multiple hate speech datasets.
\item We prepare and release LtHate, a 12k-comment Lithuanian hate speech corpus with topical and severity annotations.
\item We report a systematic comparison of six recent multilingual embedding models across Lithuanian, Russian and English hate speech corpora, with and without PCA-based compression.
\item We provide practical recommendations for model and embedding selection under computational constraints in multilingual moderation systems.
\end{itemize}

\section{Related work}
\label{sec:related}

Section overviews attempts at using multilingual text embeddings for hate speech detection in various languages.

\subsection{Hate speech and offensive language detection}

Hate speech detection has gained a lot of attention from the research community since core natural language processing techniques were established and applied. Schmidt \& Wiegand \cite{Schmidt2017} provide one of the first structured overviews of early automatic hate speech detection methods using NLP, covering classical machine learning approaches and highlighting early challenges and features (e.g., bag-of-words, lexicons). One of the more important milestones in the area of hate speech research was benchmark dataset and taxonomy introduction by Davidson et al. in 2017 \cite{Davidson2017}. Initial results were obtained by training a suite of traditional classifiers, such as logistic regression, naïve Bayes, decision trees, random forests, and support vector machines, resulting in best overall F1-score of 0.90 achieved by logistic regression while revealing a persistent confusion between hate speech and offensive content. Twitter tweets were another source for hate speech corpus \cite{Waseem2016}, which led to introducing first deep learning systems based on convolutional neural network (CNN) and multiple representations \cite{Gambaeck2017}. The best model used word2vec semantic embeddings achieved an F1-score of 78.3\% on a four-class problem, demonstrating that learned distributed representations substantially outperform surface-form features for hate speech classification. This was further confirmed in \cite{Badjatiya2017} by comparing CNN, LSTM, and FastText architectures against traditional TF-IDF and bag-of-words vector baselines on a benchmark of 16,000 annotated tweets. Their results showed that LSTM models whose embeddings were subsequently used to train gradient-boosted classifiers significantly outperformed outperformed state-of-the-art character and word n-gram methods. The introduction of transformer architecture \cite{vaswani2017} led to substantial improvements over prior sequence models as it relies exclusively on self-attention mechanisms to model dependencies between arbitrary positions in a sequence in constant time. The proposed architecture also enabled creating fine-tuned versions of the original model which usually outperformed the original. Mozafari et al. \cite{Mozafari2020a} were among the first to systematically investigate BERT fine-tuning for hate speech classification in a peer-reviewed open-access venue using BERT-Base to both Davidson \cite{Davidson2017} and Waseem and Hovy \cite{Waseem2016} datasets. By proposing a regularisation-based reweighting mechanism applied during fine-tuning, their model substantially outperformed prior deep learning baselines and helped to mitigate systematic racial biases. Further, \cite{Caselli2021} followed this direction with HateBERT pretrained on a large-scale corpus of Reddit comments containing offensive, abusive, or hateful content. Comparative experiments across three English benchmarks for abusive language detection (OffensEval, AbusEval, and HatEval) showed that HateBERT consistently outperformed the corresponding general BERT model on each of them. The OffensEval dataset \cite{Zampieri2019} has since been adopted as one of de facto standards for evaluating offensive language, and later extended to multilingual settings in OffensEval-2020 \cite{zampieri2020}, which introduced parallel datasets in Arabic, Danish, Greek, and Turkish. This led to evaluation of multilingual transformer architectures such as mBERT and XLM-RoBERTa. Across all languages, the dominant strategy among top-performing systems was fine-tuning XLM-R on the target-language training data, either alone or in combination with language-specific pre-trained models confirming the model's cross-lingual generalisation capability.

Beyond shared tasks, a growing body of work has explored the application of multilingual transformer models to heterogeneous, real-world multilingual detection settings. \cite{Roy2020} addressed the HASOC 2020 challenge on hate speech and offensive content identification in English, German, and Hindi using a two-stage hierarchical classification architecture built on mBERT and XLM-R. Their system first identified if content is hateful or offensive versus non-offensive, then classified detected toxic content into hate speech, offensive language, or profanity, exploiting the shared multilingual representation space of the underlying encoder across all three languages simultaneously. Results demonstrated that multilingual transformer encoders could successfully transfer information across typologically distinct languages within a single fine-tuned model. The authors additionally found that incorporating hashtag and emoji-aware tokenisation improved performance on Twitter data, where non-verbal signals frequently modify or intensify the meaning of hateful text. Performance of multilingual transformer models also inspired research in cross-lingual transfer as a substitute for in-language annotation in settings where labelled hate speech data is scarce or entirely absent. Bigoulaeva et al. \cite{Bigoulaeva2021}  evaluated cross-lingual transfer from high-resource source-language English to German for which only limited labelled examples were available. Using bilingual word embeddings that aligned the representation spaces of the two languages, they demonstrated that zero-shot transfer from English to German was feasible and competitive with in-language supervised baselines when training data was minimal. They also showed that using the transferred model's predictions to generate pseudo-labels for unlabelled German data and then training on those labels substantially improved over the zero-shot baseline, establishing a practical and low-cost pathway to extending detection to new languages without full annotation campaigns. However, a subsequent and more comprehensive study \cite{Bigoulaeva2022} identified structural and cultural divergence between source and target language hate speech conventions — including differences in target group definitions, slur conventions, and expression of implicit versus explicit hatred - as the primary bottleneck for effective zero-shot transfer, as well as proposed targeted data selection strategies to partially mitigate these effects. Nevertheless, further research also confirmed the capability to project knowledge from English to other languages with cross-lingual contextual embeddings, reducing dependence on language-specific labeled data and enabling transfer to low-resource settings \cite{Ranasinghe2020}, as well as outperforming general-purpose embeddings in cross-lingual classification scenarios using domain-specific multilingual hate speech embeddings \cite{Monnar2024}. Recurrent neural architectures on FastText embeddings were also adapted to multilingual hate speech detection task across English, Italian, and German languages in \cite{Corazza2020}. Awal et al. proposed HateMAML, a meta-learning framework to improve cross-lingual transfer of hate speech classifiers in low-resource languages by adapting pretrained language models to new languages with limited data. The adoption of cross-lingual transfer in low-resource scenarios has been extensively researched in the context of multiple languages, including Arabic \cite{Singhal2024}, Turkish \cite{Singhal2024}, Indian languages \cite{Singh2024, ghosh2024, chavinda2025}. Lithuanian language has only recently gained attention, with the latest study exploring transfer learning and transformer-based architectures on newly created annotated corpora of 27358 user-generated comments \cite{Mandravickaite2025}. Research results indicated that multilingual transformer models, like Multilingual BERT, LitLat BERT or Electra, can reach competitive accuracy and F1-scores.

\subsection{Multilingual transformer models}

Multilingual embeddings encode semantic information from words, sentences, paragraphs or documents across multiple languages in a shared vector space, allowing models to compare and transfer linguistic knowledge between languages. This is foundational to multilingual Natural Language Processing (NLP) tasks such as cross-lingual retrieval, semantic similarity, or machine translation. Early embedding techniques like word2vec \cite{Mikolov2013} demonstrated the power of dense vector representations for capturing semantic relationships within a single language, as well as revealed capabilities to learn semantic relationships and vector arithmetic analogies from distributional statistics of text corpora. Cross-lingual and multilingual embeddings extend this approach by representing words from multiple languages in a common space where semantically equivalent words are close together regardless of the language. Early cross-lingual methods often used linear mappings trained with bilingual dictionaries or parallel corpora to align monolingual embeddings. These mapping-based models were simple and computationally efficient, and they enabled cross-lingual lexical tasks such as bilingual lexicon induction and document classification. The early survey by Ruder et al. \cite{Ruder2017} provides a comprehensive taxonomy cross-lingual word embedding models, classifying techniques for word, sentence and document level alignments, as well as optimization objectives. These techniques were later extended to multilingual settings, handling many languages in a unified space rather than just pairs. Other researchers moved beyond mapping approaches, by developing unsupervised neural language models that jointly train on raw multilingual corpora and exploit structural similarities across languages to align representations in a common space \cite{Kanayama2017}\cite{Chen2018}. This idea was also extended to capture semantics across a wide variety of languages simultaneously, thus learning rich multilingual representations.

The rise of pre-trained multilingual transformer models completely changed the landscape of the existing multilingual embedding techniques and shifted this domain toward large, open-source transformer-based models with broad language coverage and flexible training objectives. Models like multilingual BERT (mBERT) \cite{Devlin2018} and XLM-R \cite{journalscorrabs191102116} leverage large amounts of text data from dozens or hundreds of languages to learn contextual embeddings that generalize across languages. These models are trained with self-supervised objectives (e.g., masked language modelling) on multilingual corpora and often yield strong zero-shot transfer performance. LaBSE (Language-agnostic BERT Sentence Embedding) \cite{Feng2022} extends multilingual transformer pre-training to sentence-level representations trained on parallel translation data, enabling semantically comparable sentence embeddings across more than 100 languages. Multilingual E5 models \cite{Wang2024} extend the original E5 recipe by contrastive pre-training on billions of multilingual sentence pairs followed by supervised fine-tuning and instruction tuning, yielding strong retrieval and similarity performance across many languages; the core architecture remains encoder-focused with contrastive objectives, which makes them efficient and robust but sometimes less adaptable in long-context scenarios. Qwen3-Embedding series \cite{Zhang2025} builds on the Qwen3 LLM backbone with a dense transformer architecture and multi-stage training including synthetic weak supervision, supervised fine-tuning, and model merging; these models support 100+ languages, benefitting from instruction-aware embedding generation and strong cross-lingual capabilities. Jina-Embeddings (v3/v4) use a transformer foundation with task-specific LoRA adapters and Matryoshka Representation Learning to produce flexible, high-quality embeddings for retrieval, clustering, and long-context tasks.

Approaches combining or fusing representations from multiple pre-trained models indicate that embedding choice and combination strategy significantly affect hate speech detection performance, although fusion can yield only modest gains relative to its computational cost. Other studies consider multilingual and multimodal settings, where text is combined with images and cultural context to improve detection of hateful memes and visually grounded content.

In contrast to prior work that either fine-tunes a single multilingual encoder or constructs custom domain-specific embeddings, our study compares several modern off-the-shelf sentence embedding models (Potion, Snowflake, Jina, Multilingual-E5) across multiple languages, using a fixed downstream classifier and a standardized evaluation protocol.

\section{Hate speech datasets}
\label{sec:data}

Diversity of hate speech datasets enables us to assess whether the same embedding methods and downstream machine learning models are effective across various languages and dataset sizes.

\paragraph{Lithuanian corpora - LtHate}

LtHate \cite{clarinlthate} is a new hate speech corpus for the Lithuanian language. It consists of public media comments taken from Litis \cite{clarinlitis} corpus and other public media sources. Comments from Litis corpus are sourced from two of the biggest Lithuanian online news portals in years 2010 to 2014. Comments from other media sources are spanning years 2021 to 2024 and were sourced from various social media platforms in the Lithuanian language and Lithuanian news portals.
    
The topical composition of the corpus was inspired by the methodology described in \cite{sanguinetti2018italian}. We have chosen five subjects of hate speech: a) ethnicity, nationality and race; b) gender and sexual orientation; c) country and state; d) political views and e) religion. For each subject category sets of neutral and loaded samples were collected. For each loaded sample we have noted a target of the hate and the level of the hate. The levels are 1 to 4 with 1 corresponding to expressions of contempt, 4 corresponding to outright incitement to violence against the target and 2,3 being in between. Since other corpora used in this research do not have the same structure, for the experiment described in this paper, we reduced LtHate corpus to binary labels of neutral or hate speech only and aggregated entries that represent multiple labelings of the same comment due to multiple distinct targets being present. Resulting corpus contains 5577 neutral and 6477 loaded comments in 5 categories of subjects (see Table \ref{tab:LtHate}) with a total of 12054 comments, slightly skewed towards hateful class, corresponding to 53.73\% of corpus samples.

\begin{table}[!htb]
\centering
\caption{Distribution of LtHate corpus with respect to subject categories.}
\label{tab:LtHate}
\begin{tabular}{@{}lccc@{}}
\toprule
\bf{Subject category} & \bf{Neutral} & \bf{Hate speech} & \bf{Total comments} \\
\midrule
ethnicity, nationality and race       & 2143      & 1796        & 3939             \\
gender and sexual orientation & 177       & 865         & 1042             \\
country and state               & 433       & 463         & 896              \\
political views              & 1858      & 2516        & 4374             \\
religion                        & 966       & 837         & 1803             \\
\bottomrule
\end{tabular}
\end{table}

\paragraph{Russian corpora - RuToxic.}

RuToxic is a publicly available Russian-language dataset of 163187 user comments annotated for toxicity, which has been introduced in offensive or toxic language detection research \cite{Dementieva2021}. Target class (toxic/hateful comments) here comprise 19.25\% of the dataset, therefore some weak class imbalance exist. RuToxic provides a complementary perspective on Slavic-language hate speech and toxicity, allowing us to test whether multilingual embeddings can capture similar patterns across Lithuanian and Russian data.

\paragraph{English corpora - EnSuperset.}

The English dataset EnSuperset aggregates many public hate speech and offensive language corpora into a unified binary classification benchmark \cite{Tonneau2024}, containing 360493 comments. Texts originate from English social media and online platforms, with annotations indicating presence or absence of hateful or offensive content. After harmonization of label schemes, we use a binary label 0/1 and retain only relevant textual fields ("text", "labels"). Target class encompass 27.04\% of a dataset. EnSuperset is substantially larger than LtHate and RuToxic, providing a high-resource English setting against which multilingual embeddings can be evaluated.

\section{Methodology}
\label{sec:methods}

Machine learning pipeline in Python was devised with text pre-processing, vectorization and training of models using 10-fold stratified cross-validation strategy on various hate speech datasets, and is available as open-source code at
\url{https://github.com/evavaic/KTU-Misijos-HIPSTer}. Optionally, dimensionality reduction with PCA is applied independently for each cross-validation training split.

\subsection{Text pre-processing}

All datasets investigated for hate speech detection task are processed using an identical and shared pipeline implemented in Python. Each text comment is first passed through a \emph{fix\_punctuation} function that removes exclamation marks and:

\begin{itemize}
\item normalizes encoding using \emph{ftfy} package;
\item removes hyperlinks;
\item collapses repeated punctuation marks while preserving limited emphasis;
\item standardizes spacing around punctuation and numbers, and
\item replaces emojis with shortcode text using the \emph{emoji} package.
\end{itemize}

The resulting cleaned texts are then processed with text vectorization technique to obtain feature vectors suitable for machine learning model training and testing.

\subsection{Sentence embeddings}

\begin{table}[!htb]
\centering
\caption{Description of selected multilingual embedding models. Note: MRL corresponds to Matryoshka Representation Learning, which allows truncation to lower dimensionality for resulting original feature vectors.}
\label{tab:embedding_models}
\begin{tabular}{@{}lcccl@{}}
\toprule
\bf{Embedding name} & \bf{Dimensions} & \bf{MRL} & \bf{Model size} & \bf{Model link (in huggingface.co platform)} \\
\midrule
potion    & 256        & No  & 128 M       & minishlab/potion-multilingual-128M      \\
gemma     & 768        & Yes & 308 M       & google/embeddinggemma-300m              \\
bge       & 1024       & No  & 569 M       & BAAI/bge-m3                             \\
snow      & 1024       & Yes & 568 M       & Snowflake/snowflake-arctic-embed-l-v2.0 \\
jina      & 1024       & Yes & 572 M       & jinaai/jina-embeddings-v3               \\
e5        & 1024       & No  & 560 M       & intfloat/multilingual-e5-large-instruct \\
\bottomrule
\end{tabular}
\end{table}

We compare six multilingual sentence embedding techniques for text vectorization (see Table \ref{tab:embedding_models}). Embedding models are loaded via \emph{SentenceTransformer} package, with Jina embeddings requiring \emph{trust\_remote\_code=True} setting. Texts are encoded in batches with configurable batch size, and embeddings are concatenated into a matrix with 256 dimensionality for potion, 768 for gemma and and 1024 for the remaining vectorization techniques.

\emph{minishlab/potion-multilingual-128M} \cite{minishlab2024model2vec}, \cite{huggingfaceminishlabpotionmultilingual128M} is a 0.128B parameter multilingual text embedding model distilled from BAAI/bge-m3 \cite{huggingfaceBAAIbgem3} and refined on the C4 dataset \cite{raffel2023exploringlimitstransferlearning}, \cite{huggingfaceallenaic4} by applying TokenLearn method \cite{minishlabtokenlearnblogpost}. Its notable features are high speed in comparison to larger models and unrestricted input length, due to internal averaging of embeddings of individual tokens. It has an output dimensionality of 256.

\emph{google/embeddinggemma-300m} \cite{vera2025embeddinggemmapowerfullightweighttext}, \cite{huggingfacegoogleembeddinggemma300m} is a 0.3B parameter multilingual text embedding model. The composition of dataset that the model is trained on is not disclosed, however the model does well on MMTEB \cite{enevoldsen2025mmtebmassivemultilingualtext} benchmark which includes Lithuanian language. The model uses RoPE positional encodings \cite{su2024roformer} and supports inputs up to 2048 tokens. It has an output dimensionality of 768 with matryoshka \cite{kusupati2022matryoshka} points at 512, 256 and 128 dimensions.

\emph{BAAI/bge-m3} \cite{chen2025m3embeddingmultilingualitymultifunctionalitymultigranularity}, \cite{huggingfaceBAAIbgem3} is a 0.56B parameter text embedding model specifically trained for retrieval tasks. It is trained and refined on datasets \cite{huggingfaceShitaoMLDR}, \cite{huggingfaceShitaobgem3data} that include most widespread European languages. The model supports inputs up to 8194 tokens. It has an output dimensionality of 1024.

\emph{Snowflake/snowflake-arctic-embed-l-v2.0} \cite{yu2024arcticembed20multilingualretrieval}, \cite{huggingfaceSnowflakesnowflakearcticembedlv20} is a 0.6B parameter multilingual text embedding model based on transformer encoder architecture and trained on MIRACL dataset \cite{101162tacla00595} which contains 18 languages. The model uses RoPE positional encodings \cite{su2024roformer} and supports inputs up to 8194 tokens. It has an output dimensionality of 1024, with a single matryoshka \cite{kusupati2022matryoshka} point at 256 dimensions. 

\emph{jinaai/jina-embeddings-v3} \cite{sturua2024jinaembeddingsv3multilingualembeddingstask}, \cite{huggingfacejinaaijinaembeddingsv3} is a 0.6B parameter multilingual text embedding model internally combining a transformer based encoder with 5 task specific LoRA \cite{hu2021loralowrankadaptationlarge} adapters. The model is trained on 100 languages and fine-tuned on 30 languages. Interestingly, one of those 30 languages is Latvian, which is highly similar to Lithuanian. The model uses RoPE positional encodings \cite{su2024roformer} and supports inputs up to 8194 tokens. It has an output dimensionality of 1024 with matryoshka \cite{kusupati2022matryoshka} points at 32, 64, 128, 256, 512 and 768 dimensions.

\emph{intfloat/multilingual-e5-large-instruct} \cite{wang2024multilinguale5textembeddings}, \cite{huggingfaceintfloatmultilinguale5largeinstruct} is a 0.6B parameter multilingual text embedding model based on transformer encoder architecture with weights initialized from XLM-RoBERTa large \cite{journalscorrabs191102116}, \cite{huggingfaceFacebookAIxlmrobertalarge} which is trained on 100 languages, including Lithuanian. The model then was additionally trained on datasets coming from public media and fine tuned on curated high quality datasets including one used in \cite{wang2024improving}. The model supports inputs up to 514 tokens. It has an output dimensionality of 1024.

\subsection{Dimensionality reduction}

To analyze the trade-off between performance and compactness, we train models on both original embeddings and their compressed variants. Principal Component Analysis (PCA) \cite{Hotelling1933} is a linear dimensionality reduction technique that finds a new set of orthogonal axes (principal components) that capture as much of the variance in the original feature space as possible in descending order of importance, thus providing a compressed projection of the original data.

By retaining only the first 64 principal components after PCA, we obtain a compact feature vector representation that concentrates most of the variance of the original embeddings. This reduces storage and computational costs for downstream machine learning models while also acting as a form of linear noise filtering: directions with very low variance, which often correspond to noise or redundant information, are discarded.

PCA in experiments here is fitted only to the training data of each cross-validation split to avoid information leakage and consequently the learned transformation is applied to compress both training and test embeddings within that CV split.

\subsection{Detection models}

In experiments we consider two types of downstream models for detection task:

\begin{enumerate}
    \item One-class (1c) anomaly detection. Histogram-based outlier score (HBOS) \cite{Goldstein2012} model from \emph{PyOD (Python Outlier Detection)} package, used as a one-class approach trained on the target class (hate speech) examples only. HBOS method models each feature independently using a histogram and due to its linear complexity is suitable for very large datasets, being significantly faster than many other multivariate outlier detection methods. The outlier score is estimated based on density corresponding to histogram bin each feature falls into with lower density values indicating anomalous instance. Contamination rate was fixed at 0.01, and output from model was rescaled by dividing from 10000 (for original feature vectors) or from 100 (for PCA-transformed feature vectors) to get a score resembling class probability prediction.
    \item Two-class (2c) supervised classification. Gradient boosting CatBoost classifier \cite{Prokhorenkova2018} with 500 maximum iterations, learning rate 0.05, depth 8, \emph{LogLoss} loss function, and \emph{scale\_pos\_weight} set to the ratio of negatives to positives in the training data (to address class imbalance). Within each cross-validation iteration, 80\% of the training set is used to fit the model and 20\% is held out for early stopping with a patience of 30 iterations where the best CatBoost model with respect to validation AUC ROC metric is retained.
\end{enumerate}

Although datasets had binary annotation of both target class (hate speech) and non-target class (neutral speech) examples, which is required to evaluate detection success on test folds, difference between model variants used was in training step where construction of model used data from both classes (2c case) or only data from a target class (1c case). Such selection of methods allows a direct comparison of text embedding quality under a strong two-class (2c) supervised and a weaker one-class (1c) pseudo-unsupervised setting. In practice this would correspond to the scope of annotation efforts where for the one-class case collection of only hate speech examples should be sufficient to create a detector.

\subsection{Evaluation metrics}

Machine learning experiment success was assessed using k-fold stratified cross-validation (CV) using 10 folds (k=10) and stratified by target attribute, where machine learning model is trained on all except one fold with that one fold left out to test model inference. After pooling model outputs for all test folds and comparing them to ground-truth class labels various accuracy metrics were calculated in a micro-average fashion.

To summarize detection performance, the following metrics were used:

\begin{itemize}

    \item Area under the receiver operating characteristic curve (AUC ROC), which corresponds to the probability that a randomly chosen non-target class instance will have a smaller estimated target class probability than a randomly chosen target class instance \cite{Jin_Huang_2005}. In short, AUC summarizes the probability of correctly ranking a (neutral, hate speech) pair of text examples based on detector's output and is directly related to the Wilcoxon Mann-Whitney U statistic.
    
    \item Precision-Recall curve (PRC) also allows calculating the area under the curve (AUC PRC) and for the case of large class imbalance, the PRC is recommended over ROC \cite{Beger_2016} when choosing a better performing detector.
    
    \item Overall accuracy (Accuracy) - the most known evaluation metric, calculating proportion of correct predictions to the ground-truth classes:

\begin{equation}
	    \frac {TP + TN}{TP + FP + TN + FN}
\end{equation}

where TP is true positives (correct target class predictions), TN is true negatives (correct non-target class predictions), FP is false positives (incorrect predictions of target class examples), and FN is false negatives (incorrect predictions of non-target class examples). All these counts correspond to absolute frequences from the confusion matrix, obtained after applying treshold to model's prediction (output probability of target class).

\item  Kappa \cite{cohkappa,McHugh_2012} - accuracy, corrected for class imbalance:

\begin{equation}
	    \frac {p_0 - p_e} {1 - p_e}
\end{equation}

where \emph{p\textsubscript{e}} – the sum of probabilities of predictions agreeing with the ground-truth by chance, \emph{p\textsubscript{0}} – overall accuracy of the model. According to the academic literature, the Kappa value of 0.21 - 0.40 indicates a fair agreement and 0.41 - 0.60 a moderate agreement. Higher values correspond to a substantial and almost perfect agreement result.

\end{itemize}

Plot-based AUC ROC and AUC PRC metrics are calculated using model's raw outputs before thresholding. To calculate the remaining metrics, one needs to obtain a confusion matrix by using a threshold on model's raw outputs to convert soft decision (class probability) to hard decision (class prediction). Since for the ad-hoc choice of 0.5 is usually suboptimal, we've selected a more effective - equal error rate - operating point where ROC curve intersects with diagonal and class recall metrics become equal, namely, specificity becomes approximately equal to sensitivity and, consequently, to overall accuracy.

\section{Experimental results}
\label{sec:results}

In this section we outline the main findings organized by dataset. Detailed ROC and PRC curves and accuracy metrics tables — for original feature vectors and for feature vectors after PCA transformation — are presented here with an overview of the results.

\subsection{Lithuanian dataset results}

\begin{table}[!htb]
\centering
\caption{Summary of hate speech detection success for Lithuanian language dataset.}
\label{tab:LT_results}
\begin{tabular}{@{}lcccccccc@{}}
\toprule
\multirow{2}{*}{\bf Method} & \multicolumn{2}{c}{\bf Accuracy (\%)} & \multicolumn{2}{c}{\bf Kappa} & \multicolumn{2}{c}{\bf AUC ROC} & \multicolumn{2}{c}{\bf AUC PRC}\\
 & {\bf Orig.} & {\bf PCA} & {\bf Orig.} & {\bf PCA} & {\bf Orig.} & {\bf PCA} & {\bf Orig.} & {\bf PCA}\\
\midrule
1c potion & 52.34 & 50.02 & 0.047 & 0.001 & 0.529 & 0.508 & 0.552 & 0.543 \\
1c gemma  & 54.46 & 54.11 & 0.089 & 0.082 & 0.561 & 0.560 & 0.588 & 0.580 \\
1c bge    & 56.53 & 55.13 & 0.130 & 0.102 & 0.588 & 0.574 & 0.622 & 0.596 \\
1c snow   & 57.88 & 56.41 & 0.157 & 0.128 & 0.609 & 0.588 & 0.629 & 0.605 \\
1c jina   & 63.57 & 60.52 & 0.270 & 0.209 & 0.689 & 0.645 & 0.707 & 0.655 \\
1c e5     & 60.02 & 57.64 & 0.200 & 0.152 & 0.640 & 0.605 & 0.662 & 0.623 \\
\midrule
2c potion & 75.13 & 73.92 & 0.501 & 0.477 & 0.830 & 0.818 & 0.843 & 0.831 \\
2c gemma  & 74.37 & 73.64 & 0.486 & 0.471 & 0.826 & 0.815 & 0.840 & 0.829 \\
2c bge    & 78.68 & 78.08 & 0.572 & 0.560 & 0.874 & 0.865 & 0.886 & 0.878 \\
2c snow   & 79.08 & 77.95 & 0.580 & 0.558 & 0.874 & 0.863 & 0.885 & 0.876 \\
2c jina   & 80.96 & 79.60 & 0.618 & 0.591 & 0.887 & 0.877 & 0.895 & 0.887 \\
2c e5     & 79.07 & 78.05 & 0.580 & 0.560 & 0.876 & 0.864 & 0.887 & 0.875 \\
\bottomrule
\end{tabular}
\end{table}

\begin{figure}[!htb]
\centering
\includegraphics[width=.495\linewidth]{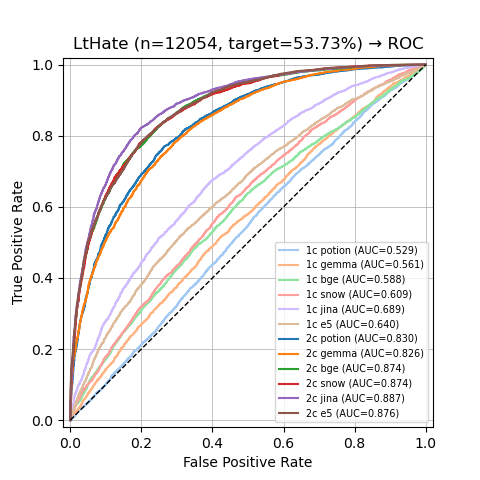}
\includegraphics[width=.495\linewidth]{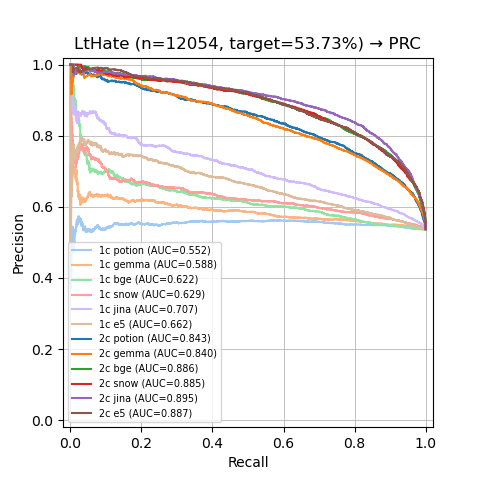}
\caption{\label{fig:LT_orig} Lithuanian language hate speech detection curves using original embeddings: ROC (left) and PRC (right).}
\end{figure}

\begin{figure}[!htb]
\centering
\includegraphics[width=.495\linewidth]{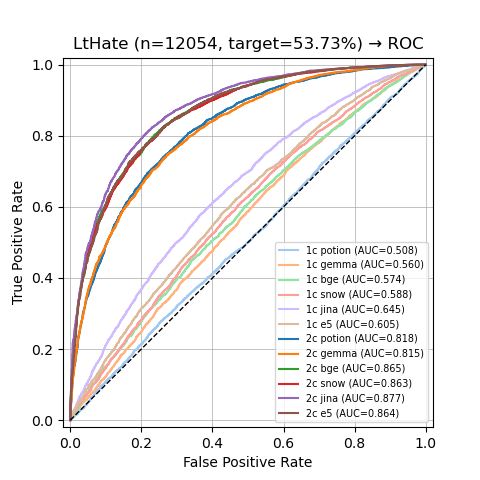}
\includegraphics[width=.495\linewidth]{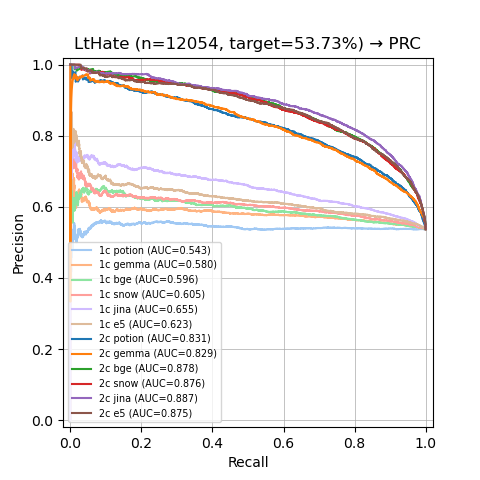}
\caption{\label{fig:LT_comp} Lithuanian hate speech detection curves using compressed embeddings: ROC (left) and PRC (right).}
\end{figure}

Machine learning for Lithuanian language hate speech dataset results are in Table \ref{tab:LT_results} and Figures \ref{fig:LT_orig}-\ref{fig:LT_comp}. One-class classification (see top part of Table \ref{tab:LT_results}) resulted in 52.34\% -- 63.57\% accuracy for original and 50.02\% -- 60.52\% accuracy for PCA-compressed embeddings. Two-class classification (see bottom part of Table \ref{tab:LT_results}) resulted in 74.37\% -- 80.96\% accuracy for original and 73.64\% -- 79.60\% accuracy for PCA-compressed embeddings. Two-class supervised classification clearly outperformed one-class anomaly detection with only a negligible differences between original and PCA-compressed embeddings in the two-class case.

Detection effectiveness for Lithuanian hate speech, as measured by ROC/PRC curves and accuracy metrics, was highest for Jina embeddings, resulting in AUC ROC of 0.887 and AUC PRC of 0.895 for two-class classification (see Fig. \ref{fig:LT_orig}) and PCA transformation did not affect this result noticeably (see Fig. \ref{fig:LT_comp}). Other very competitive embeddings in two-class case were bge, snow and e5. However, gemma embeddings demonstrated surprisingly bad result and even were slightly outperformed by lower dimensional potion embeddings in two-class case.

Slightly above moderate agreement between predicted class and ground truth (best Kappa=0.62 for jina embeddings) demonstrates average success in hate speech detection for Lithuanian language dataset.

\subsection{Russian dataset results}

\begin{table}[!htb]
\centering
\caption{Summary of hate speech detection success for Russian language dataset.}
\label{tab:RU_results}
\begin{tabular}{@{}lcccccccc@{}}
\toprule
\multirow{2}{*}{\bf Method} & \multicolumn{2}{c}{\bf Accuracy (\%)} & \multicolumn{2}{c}{\bf Kappa} & \multicolumn{2}{c}{\bf AUC ROC} & \multicolumn{2}{c}{\bf AUC PRC}\\
 & {\bf Orig.} & {\bf PCA} & {\bf Orig.} & {\bf PCA} & {\bf Orig.} & {\bf PCA} & {\bf Orig.} & {\bf PCA}\\
\midrule
1c potion & 67.71 & 59.40 & 0.254 & 0.126 & 0.741 & 0.626 & 0.422 & 0.290 \\
1c gemma  & 79.00 & 70.87 & 0.462 & 0.308 & 0.869 & 0.776 & 0.666 & 0.462 \\
1c bge    & 74.50 & 67.87 & 0.374 & 0.257 & 0.816 & 0.741 & 0.569 & 0.436 \\
1c snow   & 69.18 & 66.12 & 0.279 & 0.228 & 0.767 & 0.720 & 0.467 & 0.380 \\
1c jina   & 76.15 & 69.87 & 0.405 & 0.291 & 0.841 & 0.763 & 0.622 & 0.445 \\
1c e5     & 80.82 & 75.28 & 0.500 & 0.389 & 0.890 & 0.831 & 0.712 & 0.579 \\
\midrule
2c potion & 85.41 & 84.97 & 0.601 & 0.591 & 0.934 & 0.930 & 0.814 & 0.802 \\
2c gemma  & 89.41 & 88.73 & 0.698 & 0.681 & 0.961 & 0.957 & 0.885 & 0.873 \\
2c bge    & 90.99 & 90.33 & 0.739 & 0.722 & 0.970 & 0.967 & 0.906 & 0.895 \\
2c snow   & 91.04 & 90.52 & 0.740 & 0.727 & 0.971 & 0.968 & 0.912 & 0.903 \\
2c jina   & 91.07 & 90.62 & 0.741 & 0.729 & 0.972 & 0.969 & 0.909 & 0.901 \\
2c e5     & 92.19 & 91.74 & 0.770 & 0.759 & 0.978 & 0.975 & 0.931 & 0.924 \\
\bottomrule
\end{tabular}
\end{table}

\begin{figure}[!htb]
\centering
\includegraphics[width=.495\linewidth]{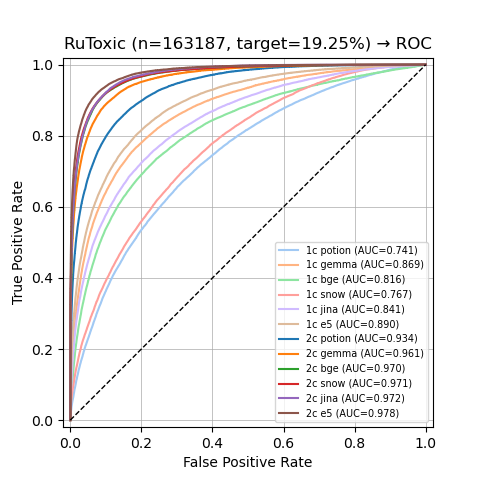}
\includegraphics[width=.495\linewidth]{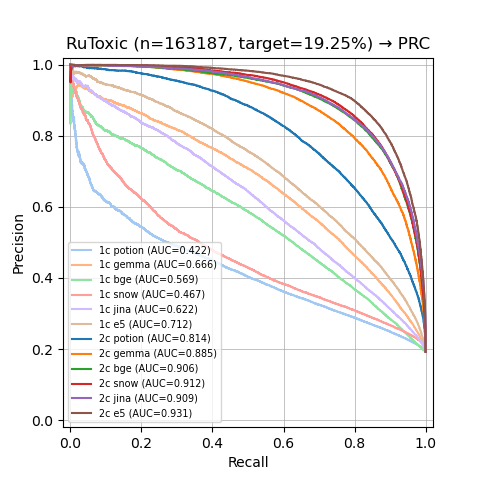}
\caption{\label{fig:RU_orig} Russian hate speech detection curves using original embeddings: ROC (left) and PRC (right).}
\end{figure}

\begin{figure}[!htb]
\centering
\includegraphics[width=.495\linewidth]{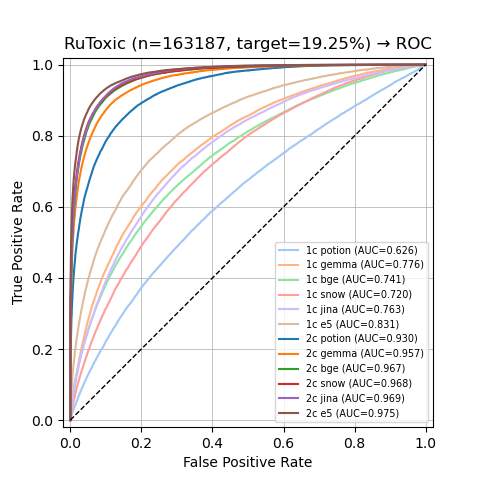}
\includegraphics[width=.495\linewidth]{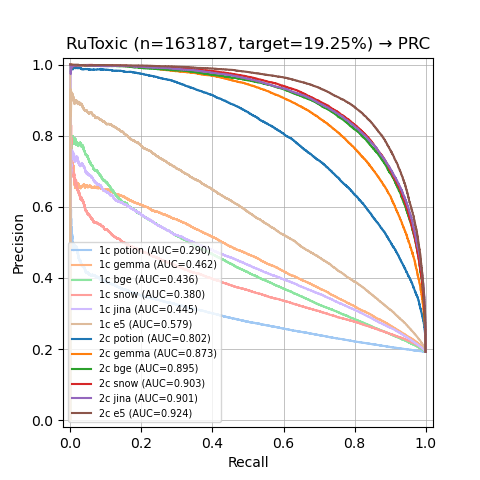}
\caption{\label{fig:RU_comp} Russian language hate speech detection curves using compressed embeddings: ROC (left) and PRC (right).}
\end{figure}

Machine learning for Russian language hate speech dataset results are in Table \ref{tab:RU_results} and Figures \ref{fig:RU_orig}-\ref{fig:RU_comp}. One-class classification (see top part of Table \ref{tab:RU_results}) resulted in 67.71\% -- 80.82\% accuracy for original and 59.40\% -- 75.28\% accuracy for PCA-compressed embeddings. Two-class classification (see bottom part of Table \ref{tab:RU_results}) resulted in 85.41\% -- 92.19\% accuracy for original and 84.97\% -- 91.74\% accuracy for PCA-compressed embeddings. Two-class supervised classification clearly outperformed one-class anomaly detection with only a negligible differences between original and PCA-compressed embeddings in two-class case.

Detection effectiveness for Russian hate speech, as measured by ROC/PRC curves and accuracy metrics, was highest for E5 embeddings, resulting in AUC ROC of 0.978 and AUC PRC of 0.931 for two-class classification (see Fig. \ref{fig:RU_orig}) and PCA transformation did not affect it noticeably (see Fig. \ref{fig:RU_comp}). Other embeddings (jina, snow, bge) also were very competitive to e5 in two-class case, with the gemma embedding performing only slightly worse. Worst performance was for snow and potion embeddings.

Substantial agreement between predicted class and ground truth (best Kappa=0.77 for e5 embeddings) demonstrates high success in hate speech detection for Russian language dataset.

\subsection{English dataset results}

\begin{table}[!htb]
\centering
\caption{Summary of hate speech detection success for English language dataset.}
\label{tab:EN_results}
\begin{tabular}{@{}lcccccccc@{}}
\toprule
\multirow{2}{*}{\bf Method} & \multicolumn{2}{c}{\bf Accuracy (\%)} & \multicolumn{2}{c}{\bf Kappa} & \multicolumn{2}{c}{\bf AUC ROC} & \multicolumn{2}{c}{\bf AUC PRC}\\
 & {\bf Orig.} & {\bf PCA} & {\bf Orig.} & {\bf PCA} & {\bf Orig.} & {\bf PCA} & {\bf Orig.} & {\bf PCA}\\
\midrule
1c potion & 55.27 & 51.32 & 0.085 & 0.021 & 0.572 & 0.516 & 0.319 & 0.289 \\
1c gemma  & 66.69 & 58.01 & 0.283 & 0.131 & 0.731 & 0.613 & 0.484 & 0.376 \\
1c bge    & 60.25 & 52.03 & 0.169 & 0.032 & 0.647 & 0.526 & 0.379 & 0.291 \\
1c snow   & 60.98 & 55.14 & 0.182 & 0.083 & 0.663 & 0.575 & 0.417 & 0.328 \\
1c jina   & 61.94 & 55.84 & 0.198 & 0.094 & 0.671 & 0.583 & 0.416 & 0.343 \\
1c e5     & 64.22 & 56.63 & 0.239 & 0.108 & 0.702 & 0.593 & 0.464 & 0.338 \\
\midrule
2c potion & 70.14 & 69.55 & 0.347 & 0.336 & 0.777 & 0.770 & 0.570 & 0.558 \\
2c gemma  & 76.37 & 76.67 & 0.468 & 0.474 & 0.848 & 0.851 & 0.679 & 0.683 \\
2c bge    & 74.50 & 74.10 & 0.431 & 0.423 & 0.828 & 0.824 & 0.650 & 0.644 \\
2c snow   & 75.89 & 76.00 & 0.459 & 0.461 & 0.842 & 0.844 & 0.668 & 0.670 \\
2c jina   & 75.35 & 75.02 & 0.448 & 0.442 & 0.837 & 0.834 & 0.657 & 0.655 \\
2c e5     & 76.95 & 77.21 & 0.480 & 0.485 & 0.855 & 0.859 & 0.698 & 0.705 \\
\bottomrule
\end{tabular}
\end{table}

\begin{figure}[!htb]
\centering
\includegraphics[width=.495\linewidth]{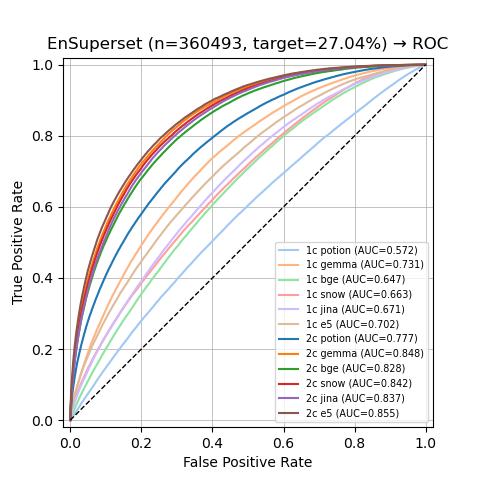}
\includegraphics[width=.495\linewidth]{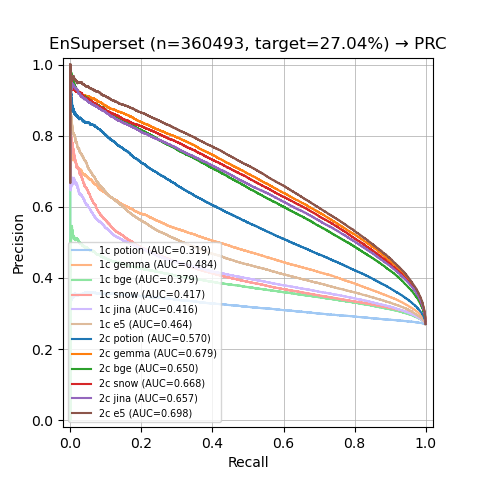}
\caption{\label{fig:EN_orig} English language hate speech detection curves using original embeddings: ROC (left) and PRC (right).}
\end{figure}

\begin{figure}[!htb]
\centering
\includegraphics[width=.495\linewidth]{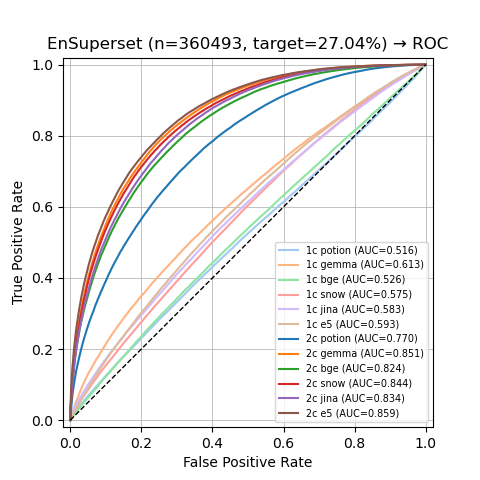}
\includegraphics[width=.495\linewidth]{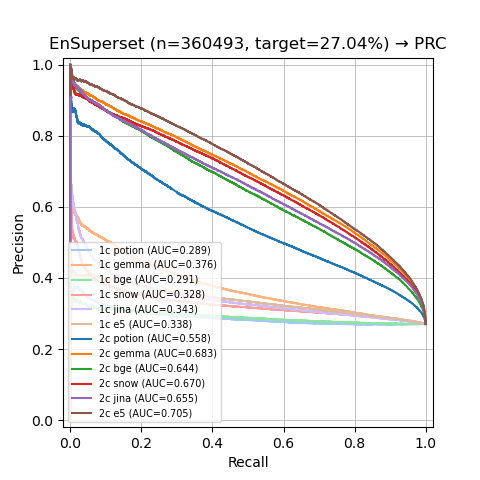}
\caption{\label{fig:EN_comp} English language hate speech detection curves using compressed embeddings: ROC (left) and PRC (right).}
\end{figure}

Machine learning for English language hate speech dataset results are in Table \ref{tab:EN_results} and Figures \ref{fig:EN_orig}-\ref{fig:EN_comp}. One-class classification (see top part of Table \ref{tab:EN_results}) resulted in 55.27\% -- 64.22\% accuracy for original and 51.32\% -- 56.63\% accuracy for PCA-compressed embeddings. Two-class classification (see bottom part of Table \ref{tab:EN_results}) resulted in 70.14\% -- 76.95\% accuracy for original and 69.55\% -- 77.21\% accuracy for PCA-compressed embeddings (see bottom part of Table \ref{tab:EN_results}). Two-class supervised classification clearly outperformed one-class anomaly detection with only a negligible differences between original and PCA-compressed embeddings in two-class case.

Detection effectiveness for English hate speech, as measured by ROC/PRC curves and accuracy metrics, was highest for E5 embeddings, resulting in AUC ROC of 0.855 and AUC PRC of 0.698 for two-class classification (see Fig. \ref{fig:EN_orig}) and PCA transformation even improved this result slightly (see Fig. \ref{fig:EN_comp}). Other very competitive embeddings in two-class case were gemma, snow and jina, with surprisingly good results for lower-dimensional gemma embeddings.

Moderate agreement between predicted class and ground truth (best Kappa=0.48 for e5 embeddings) demonstrates below average success in hate speech detection for English language dataset.

\subsection{Overview of all results}

Across the three hate speech datasets and six multilingual embedding models, several consistent patterns can be observed. First, two-class (2c) supervised CatBoost classifiers systematically and substantially outperform one-class (1c) HBOS anomaly detectors in terms of accuracy, Kappa, AUC ROC, and AUC PRC for all languages and embedding families. For example, on Lithuanian LtHate, accuracy increases from 63.57\% (best 1c, jina) to 80.96\% (best 2c, jina), while on Russian RuToxic it increases from 80.82\% (best 1c, e5) to 92.19\% (best 2c, e5) and on English EnSuperset from 66.69\% (best 1c, gemma) to 76.95\% (best 2c, e5). These results suggest that even when high-quality negative examples are more expensive to annotate strong supervised approach should be preferred whenever a reasonably balanced labeled dataset can be constructed.

Second, detection effectiveness depends strongly on both language and embedding model. For Lithuanian LtHate, the best performing configuration combines Jina embeddings with the 2c CatBoost classifier, reaching accuracy of 80.96\%, Kappa of 0.618, AUC ROC of 0.887, and AUC PRC of 0.895. Russian RuToxic achieves the highest overall scores in the study, with e5 + 2c CatBoost attaining 92.19\% accuracy, Kappa of 0.770, AUC ROC of 0.978, and AUC PRC of 0.931. For English EnSuperset, e5 again provides the strongest results in the 2c setting with 76.95\% accuracy, Kappa of 0.480, AUC ROC of 0.855, and AUC PRC of 0.698, slightly improving further when PCA-compressed embeddings are used. Overall, modern large multilingual encoders (jina, e5, snow, bge, gemma) consistently outperform the compact potion model, although potion remains surprisingly competitive given its much smaller dimensionality and parameter count.

Third, the relative ranking of embedding models is not fully consistent across languages, reflecting differences in training data coverage and linguistic similarity. Jina embeddings are clearly superior for Lithuanian LtHate in the 2c setting, whereas e5 dominates on both Russian RuToxic and English EnSuperset. Gemma embeddings behave somewhat atypically, yielding only mid-tier performance on Lithuanian (even slightly worse than potion in the 2c case) while becoming one of the top-performing models on English EnSuperset and achieving strong 1c performance on RuToxic. This observation highlights that embedding models with excellent global multilingual benchmarks do not necessarily transfer uniformly across all lower-resource target languages and domains.

Fourth, PCA-based dimensionality reduction to 64 components preserves most of the discriminative information for the 2c CatBoost classifiers. Across all datasets and embedding models, accuracy, Kappa, AUC ROC, and AUC PRC values for original and PCA-compressed representations differ only marginally in the 2c setting, often within one percentage point. For example, for Lithuanian LtHate with jina embeddings, AUC ROC drops only from 0.887 to 0.877 and AUC PRC from 0.895 to 0.887, while for Russian RuToxic with e5 the AUC ROC remains at 0.978 vs. 0.975 and AUC PRC at 0.931 vs. 0.924. However, in the 1c HBOS scenario PCA compression can noticeably degrade performance, particularly for RuToxic and EnSuperset, indicating that fine-grained density information is more important for histogram-based anomaly scoring than for gradient-boosted decision trees.

Finally, there are clear differences in achievable performance across languages and datasets. Russian RuToxic, which is relatively large and has a moderate class imbalance, yields the highest scores (best Kappa = 0.770), suggesting that current multilingual embeddings can model Russian toxic language patterns particularly well under a supervised 2c setup. Lithuanian LtHate attains lower but still competitive results (best Kappa = 0.618), reflecting both its smaller size and the increased difficulty of modeling a newly constructed low-resource language hate speech corpus. English EnSuperset yields intermediate performance (best Kappa = 0.485 with PCA-compressed e5 embeddings in 2c case), which is slightly lower than might be expected for English but may be explained by the heterogeneity of source corpora and label schemes that were harmonized into a single binary benchmark.

\section{Discussion \& Conclusions}

In this paper, we presented a comparative study of six modern multilingual sentence embedding models — potion, gemma, bge, snow, jina, and e5 — for hate speech detection in Lithuanian, Russian, and English. We introduced LtHate, a new Lithuanian hate speech corpus with detailed topical and severity annotations that we reduced to a binary classification setting for experiments, and we evaluated all embedding models in both one-class (HBOS) and two-class (CatBoost) configurations with and without PCA-based dimensionality reduction. Experimental results show that contemporary multilingual encoders combined with a simple gradient-boosted classifier can achieve moderate to substantial agreement with human annotations across all three languages, with the strongest performance observed for Russian and competitive results for the newly created Lithuanian dataset.

From a practical perspective, the experiments suggest several recommendations for multilingual hate speech detection systems. Whenever it is feasible to obtain labeled examples for both hateful and non-hateful categories, a two-class supervised setup with CatBoost (or similar gradient-boosting methods) should be preferred over purely one-class anomaly detection, as the latter consistently lags behind across datasets and metrics. Among the embedding models, jina embeddings appear to be the most suitable choice for Lithuanian LtHate, whereas e5 embeddings provide the best overall performance for Russian RuToxic and English EnSuperset; other embeddings (snow, bge, gemma) remain competitive alternatives. Additionally, our findings indicate that applying PCA to reduce embeddings to 64 principal components yields almost no loss in 2c classification performance, offering a straightforward way to lower memory and computation costs in real-world deployments.

At the same time, several limitations of the current study point to directions for future work. First, we focused exclusively on off-the-shelf sentence encoders without any task-specific fine-tuning, meaning that further gains are likely achievable via supervised or contrastive adaptation on in-domain hate speech corpora. Second, our experiments considered only binary hate vs. neutral labels, whereas LtHate and many existing datasets provide richer taxonomies (e.g., fine-grained target groups, severity levels), and modeling these distinctions may be necessary for more nuanced moderation decisions. Third, the current setup is text-only and does not incorporate multi-modal information such as images, emojis beyond textual normalisation, or conversation context, which are often crucial in real-world hateful or abusive content.

Therefore, an important direction for further research would be systematic evaluation of instruction-tuned large language models in zero-shot and few-shot classification regimes, as well as hybrid architectures combining frozen multilingual embedding encoders with lightweight adapters fine-tuned on hate speech and toxicity detection tasks. Also, integrating explainability techniques and bias assessments into the evaluation protocol will be essential for understanding and mitigating potential harms when deploying multilingual hate speech detectors in high-stakes, real-world moderation scenarios.

\subsection*{Code availability}

All code used to implement the unified Python pipeline for pre-processing, multilingual sentence embeddings, PCA compression, and model training is publicly available at
\url{https://github.com/evavaic/KTU-Misijos-HIPSTer}.

\section*{Acknowledgements}

This work was conducted as part of the execution of the project "Mission-driven Implementation of Science and Innovation Programs" (No. 02-002-P-0001), funded by the Economic Revitalization and Resilience Enhancement Plan "New Generation Lithuania".

\bibliographystyle{unsrtnat}
\bibliography{references}

\end{document}